%% file: main.tex
\RequirePackage[T1]{fontenc}
\documentclass[letterpaper,10pt,conference]{ieeeconf}

\IEEEoverridecommandlockouts
\usepackage{graphicx}
\usepackage{amsmath,amssymb}
\usepackage{booktabs}
\usepackage{multirow}
\usepackage{array}
\usepackage{cite}
\usepackage{xcolor}
\usepackage{url}

\input{config}

\input{macros}

\title{\LARGE \bfseries
\papertitle
}

\ifanonymous
\author{Anonymous Authors}
\else
\author{
Jiangong Xiao$^{1,2}$,
Zhe Sun$^{1,2,*}$,
Kanzhong Yao$^{2}$,
Yuanbo Bi$^{2}$,
Haofei Zhao$^{2}$,
Ruixuan Hu$^{2}$,
Guan Huang$^{1}$,\\
and Xuelong Li$^{2}$%
\thanks{$^{1}$Northwestern Polytechnical University, Xi'an 710072, China.}%
\thanks{$^{2}$Institute of Artificial Intelligence (TeleAI), China Telecom, Shanghai 200232, China.}%
\thanks{Work done during an internship at TeleAI.}%
\thanks{$^{*}$Corresponding author: Zhe Sun (sunzhe@nwpu.edu.cn).}%
}
\fi

\begin{document}
\raggedbottom

\maketitle
\thispagestyle{empty}
\pagestyle{empty}

\begin{abstract}
\input{sections/00_abstract}
\end{abstract}

\input{sections/01_introduction}
\input{sections/02_related_work}
\input{sections/03_problem_formulation}
\input{sections/04_ground_truth}
\input{sections/05_tracking_method}

\input{sections/06_experiments}
\input{sections/07_conclusion}

\bibliographystyle{IEEEtran}
\bibliography{references}

\end{document}

%% file: config.tex
\newif\ifanonymous
\anonymousfalse

\newcommand{\papertitle}{Continuous Ground-Truth Construction and a Recovery Policy for Air--Water Robotic Tracking}
\newcommand{\methodname}{\textsc{CMRP}}

\newcommand{\crossframes}{22,346}
\newcommand{\crosssequences}{18}

\newcommand{\projectionpoints}{14,984}

%% file: macros.tex
\newcommand{\best}[1]{\textbf{#1}}

\renewcommand{\arraystretch}{1.08}

%% file: sections/00_abstract.tex
Visual tracking across the air--water interface is challenged by splashes, bubbles, refraction, reflections, and abrupt appearance changes that can temporarily invalidate observations. This setting poses two coupled difficulties: 1) for evaluation, image-only annotation cannot reliably describe the target's physical location during visual blindness; 2) for online tracking, corrupted observations can contaminate motion estimates and appearance templates. We address the first difficulty with a construction pipeline that synchronizes camera frames with motion-capture poses, projects known target geometry, corrects underwater projection with a medium-gated residual, and subjects the annotations to manual review. This yields an evaluation-only cross-medium test set of {\crossframes} frames. We further introduce a Cross-Medium Recovery Policy ({\methodname}) centered on confidence-triggered template selection. It supplies MixFormerV2 with the fixed initial template, a window-best pre-trigger template, and a trigger-frame Kalman-guided image crop, together with their associated weights, without retraining the visual backbone. In the accuracy evaluation, {\methodname} achieves $49.90$ Macro Success AUC, $2.95$ points above MixFormerV2 Official.On selected cross-medium transition and occlusion--recovery
intervals, {\methodname} increases MixFormerV2 tracking coverage
from $47.91\%$ to $50.43\%$ relative to Official updating,
while mean loss-to-recovery latency over successfully recovered
videos decreases from $55.3$ to $49.3$ frames.

\vspace{0.4em}
\noindent\textbf{Index Terms---}cross-medium object tracking, air--water robotics, motion capture, ground-truth construction, target recovery, template management.

%% file: sections/01_introduction.tex
\section{Introduction}

\begin{figure}[!t]
  \centering
  \includegraphics[width=\columnwidth]{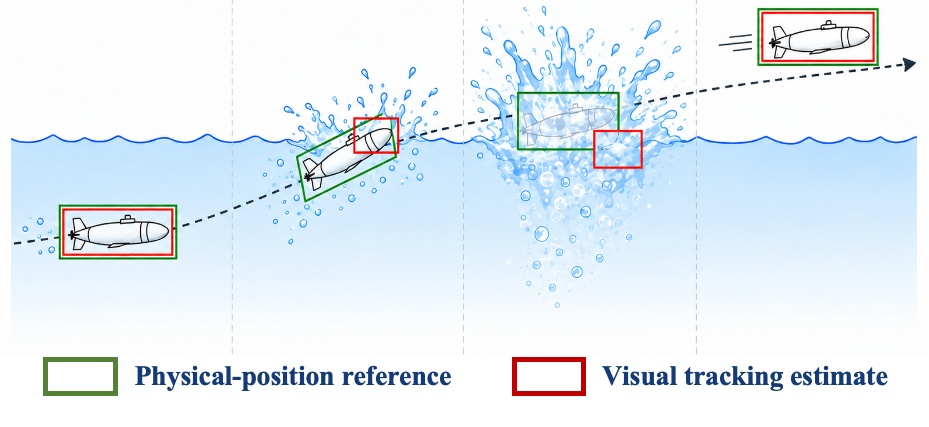}
  \caption{Schematic of tracking during water exit. Green boxes indicate the physical-position reference used only for offline evaluation; red boxes illustrate visual tracking estimates. Splashes and partial visibility can cause localization errors even as the target continues moving, motivating recovery from image observations alone.}
  \label{fig:introduction}
\end{figure}

Amphibious air--water robots offer opportunities for environmental monitoring, search and rescue, and operations spanning both media~\cite{siddall2017aquamav,chen2017aerialaquatic,li2022aerialaquatic}. In collaborative applications, one platform may need to continuously observe another target while it approaches, crosses, or leaves the water surface. At the interface, splashes, bubbles, reflections, exposure changes, and refraction can abruptly distort the target or make it entirely invisible for several frames. As illustrated in Fig.~\ref{fig:introduction}, the target remains physically present and continues to move even when interface-induced disturbances temporarily invalidate visual observations.

This setting first creates an evaluation problem. When the target is only partially visible, manual annotation can be ambiguous; when splashes or bubbles fully obscure it, the image alone may not reliably reveal its physical position. Excluding these frames from localization evaluation leaves the interval in which drift accumulates unmeasured. Cross-medium evaluation therefore benefits from a reference location that remains defined independently of instantaneous visibility. A natural question arises: is this process measurable?

To answer this, we use motion capture during offline ground-truth construction. Since the target itself is not directly observable to the motion-capture system throughout the crossing, we rigidly attach an extended marker structure to the target and track its pose instead. Camera frames and target poses are synchronized by a shared timestamp, and a calibrated transformation chain projects a three-dimensional reference point and known target geometry into the image. A medium-gated residual corrects systematic underwater displacement, after which every generated sequence is manually inspected and corrected. This procedure yields an evaluation-only test set of {\crossframes} frames with continuous physical-position boxes.
This procedure yields an evaluation-only test set of
{\crossframes} frames with continuous physical-position
boxes.\footnote{The evaluation dataset and code will be
made publicly available upon acceptance.}

The same visual degradation also creates a tracking problem. Unreliable visual observations can corrupt online template updates and propagate tracking errors~\cite{zhang2019updatenet,an2023trusted,yang2026samurai}. Near the interface, a mistaken target response can move the next search region and overwrite useful appearance information. Earlier clear observations preserve identity but may poorly represent the changing appearance, whereas repeated updates during severe degradation risk contamination. This motivates selecting recent references at the onset of low confidence while retaining a fixed identity anchor.

For online tracking, we propose a Cross-Medium Recovery Policy ({\methodname}) built on MixFormerV2~\cite{cui2023mixformerv2} without retraining its visual backbone. At the onset of low confidence, it selects the highest-confidence template from the preceding window and an image crop at the trigger-frame Kalman prediction, while preserving the initial template. The selected references are retained during sustained low confidence, with motion prediction and recovery search providing position support. Fig.~\ref{fig:overview} summarizes the offline ground-truth construction and online tracking framework.

Our main contributions are as follows:
\begin{enumerate}
  \item We introduce a motion-capture-based procedure for constructing continuous air--water tracking ground truth. Synchronized pose projection, medium-gated residual correction, known target geometry, and manual review provide boxes even when the image observation is unreliable.
\item We establish an {\crosssequences}-video, {\crossframes}-frame evaluation-only cross-medium test set and define protocols for standard tracking accuracy,
frame-level tracking coverage, and video-level recovery latency.
  \item We propose {\methodname}, a confidence-triggered template policy combining a fixed identity reference, a pre-trigger window-best template, and a trigger-frame Kalman-guided crop. Experiments evaluate its cross-medium accuracy and recovery behavior.
\end{enumerate}

%% file: sections/02_related_work.tex
\section{Related Work}

\subsection{Cross-Medium Robotic Tracking}

Aerial--aquatic robots demonstrate several approaches to crossing the interface~\cite{alzu2018looncopter,mercado2019aerialunderwater,zufferey2019jumpgliding}.
AquaMAV and Dipper use reconfigurable wings for air--water transitions~\cite{siddall2017aquamav,rockenbauer2021dipper};
flapping-wing robots combine aerial and aquatic locomotion with water-exit mechanisms~\cite{chen2017aerialaquatic,zufferey2026leaping};
and Li et al.\ demonstrate a self-contained platform with morphing propellers
and surface adhesion~\cite{li2022aerialaquatic}.
Recent work has also explored user-oriented multimodal FPV drone systems
for aquatic-aerial operation~\cite{bi2026useroriented}.
These studies address mobility and physical transition constraints.
Our work complements them by constructing continuous target annotations and evaluating visual tracking through interface-induced observation degradation.

\subsection{Visual Object Tracking}
Transformer-based single-object trackers have substantially improved the modeling of target-template and search-region interactions. OSTrack~\cite{ye2022ostrack} unifies feature extraction and target-search relation modeling in a one-stream Transformer. MixFormer~\cite{cui2022mixformer} uses mixed attention to exchange information between the template and search region. MixFormerV2~\cite{cui2023mixformerv2} further reduces computational complexity through prediction tokens and model distillation. ODTrack~\cite{zheng2024odtrack} associates target information across consecutive frames by propagating temporal tokens online.

Template adaptation is also well studied. UpdateNet learns to combine initial, accumulated, and current appearances~\cite{zhang2019updatenet}; local trusted templates select reliable history within a sliding window~\cite{an2023trusted}; and SAMURAI combines Kalman motion prediction with memory selection~\cite{yang2026samurai}. Long-term tracking research studies target disappearance, re-detection, and update control~\cite{valmadre2018oxuva,lukezic2021longterm,dai2020ltmu}. Our focus is the effectiveness of a low-confidence-triggered policy for cross-medium tracking: selecting pre-trigger appearance and a trigger-frame motion-guided crop, then retaining them during sustained low confidence.

\subsection{Underwater Perception and Cross-Medium Tracking Data}
Underwater imaging is affected by wavelength-dependent attenuation, backscatter, and range-dependent color changes~\cite{akkaynak2019seathru}. Cross-medium sequences additionally contain splashes, reflections, bubbles, partial entry, and abrupt exposure changes near the interface. Conventional benchmarks such as LaSOT~\cite{fan2019lasot} and UAV123~\cite{mueller2016uav123} provide standard tracking evaluations. Underwater-specific benchmarks such as UTB180 and WebUOT-1M evaluate visual tracking under underwater appearance degradation~\cite{alawode2023utb180,zhang2024webuot1m}. Here, however, a continuous physical-position reference is needed to evaluate localization even when the target is obscured. We therefore use synchronized motion-capture projection to construct labels independently of the online tracker.

%% file: sections/03_problem_formulation.tex
\section{Problem Formulation and Evaluation}
\label{sec:problem}

Given an image sequence $\{I_t\}_{t=1}^{T}$ and the initial target box $B_1$, an online tracker predicts a box $B_t$ in each frame, and $q_t$ is the confidence directly output by the model. The offline construction branch provides a ground-truth box $B_t^*$ even when splashes or bubbles obscure the target. The online {\methodname} branch receives only the initial box and image stream; it has no access to motion-capture poses, medium labels, or future frames.

\begin{figure*}[t]
  \centering

  {\IfFileExists{figures/pipline.png}{%
    \includegraphics[
      width=0.9\textwidth,
      height=0.8\textwidth
    ]{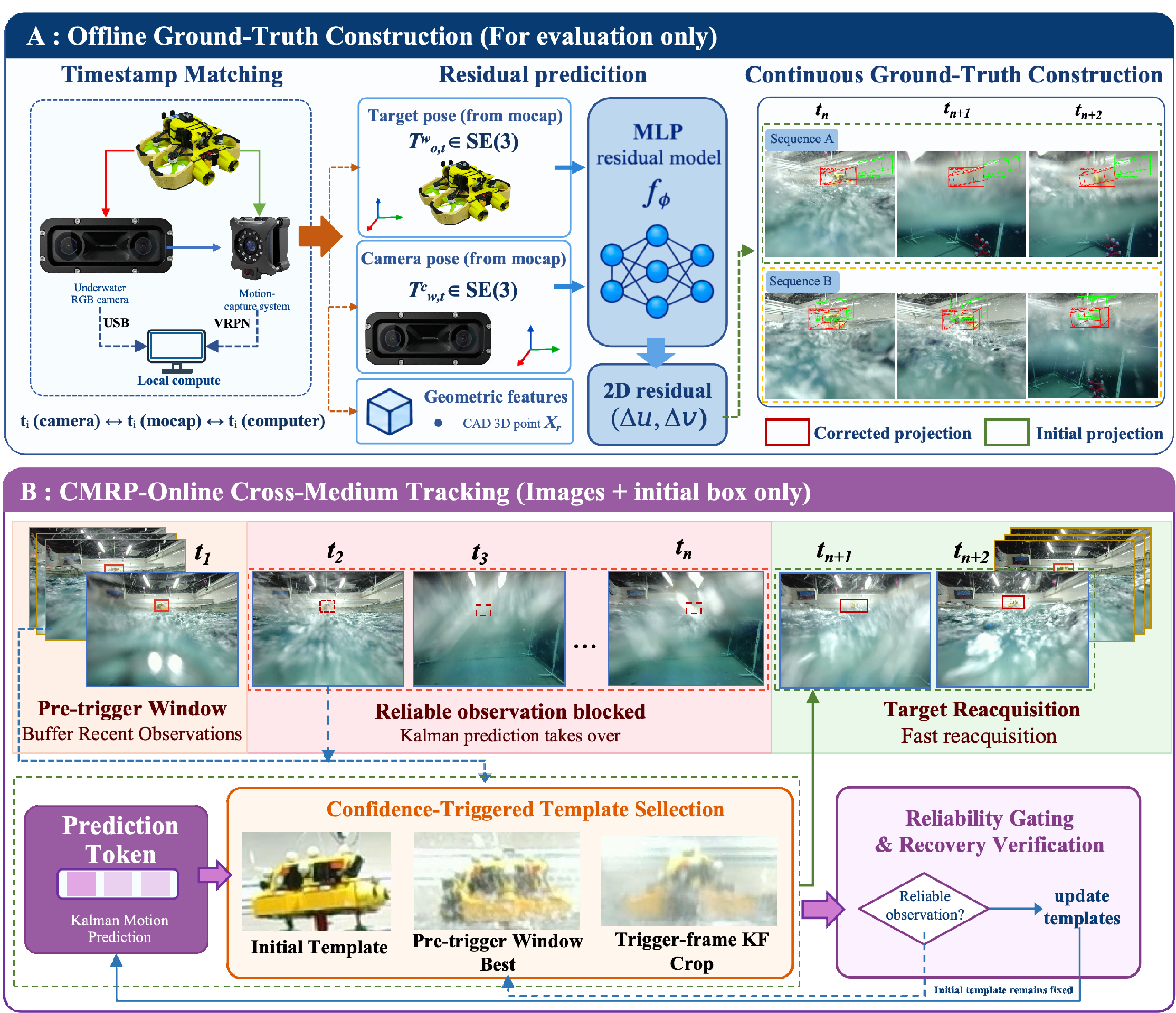}%
  }{%
    {\color{gray!40}\fbox{\rule{0pt}{58mm}\rule{0.97\textwidth}{0pt}}}%
  }}
  \caption{Offline annotation and online tracking. The offline branch constructs continuous labels from synchronized motion capture. The online branch uses confidence-triggered template selection and motion guidance, without access to motion-capture measurements.}
  \label{fig:overview}
\end{figure*}

\subsection{Standard Tracking Accuracy}
Following standard tracking evaluation~\cite{wu2013otb}, the success rate $S_s(\tau)$ for sequence $s$ is the fraction of frames whose IoU reaches threshold $\tau$. Success AUC summarizes this curve over $\tau\in[0,1]$; its continuous integral equals sequence mean IoU, whereas a sampled curve gives a discrete approximation. We average sequence-level AUC values with equal weight and denote the result as Macro Success AUC. Precision, normalized precision, and overlap precision at IoU thresholds $0.50$ and $0.75$ are retained as complementary metrics.

\subsection{Tracking Coverage and Recovery}

Tracking coverage is evaluated on 9,166 frames from selected
cross-medium transition and occlusion--recovery intervals
in the test set. It is defined as the percentage of these
frames satisfying $\operatorname{IoU}(B_t,B_t^*)>0.5$.

Recovery latency is evaluated on the corresponding videos,
with one recovery outcome per video.
All ground-truth comparisons are used only for offline evaluation.
For video $s$, timing starts at the first frame
$t_s^{\mathrm{lost}}$ satisfying
\begin{equation}
  \operatorname{IoU}(B_t,B_t^*) < 0.15.
  \label{eq:lost}
\end{equation}

Stable recovery requires five consecutive frames in which
both the predicted-box overlap and the tracker-reported
confidence satisfy
\begin{equation}
  \operatorname{IoU}(B_t,B_t^*) > 0.5,
  \qquad q_t > 0.5.
  \label{eq:recovered}
\end{equation}

After this run is confirmed, $t_s^{\mathrm{rec}}$ is assigned
to its first frame. We use the first qualifying run after
$t_s^{\mathrm{lost}}$ and record
$L_s=t_s^{\mathrm{rec}}-t_s^{\mathrm{lost}}$;
later recoveries in the same video are not additional trials.
Let $\mathcal R$ contain the successfully recovered videos.
Mean recovery latency is
$\overline L=\frac{1}{|\mathcal R|}\sum_{s\in\mathcal R}L_s$.
Videos without recovery are excluded from the latency mean.
Timing starts at tracking loss and can include part of the
invisible interval.

%% file: sections/04_ground_truth.tex
\section{Continuous Ground Truth and Test-Set Construction}

Our objective is to retain a physically grounded target location when image observations are unreliable.
Motion capture is used only as an offline measurement instrument for label construction and is never available to the evaluated trackers.
Motion capture provides independent pose ground truth in robotic benchmarks~\cite{burri2016euroc}.
Prior datasets address temporal alignment~\cite{schubert2018tumvi} and marker-to-camera calibration~\cite{mueggler2017eventdataset}.
EV-IMO combines tracked poses with scanned geometry to generate depth and segmentation labels~\cite{mitrokhin2019evimo}. 

\subsection{Acquisition Setup and Pose Measurement}
\label{sec:mocap_setup}

To continuously measure camera and target poses during
air--water transitions, we rigidly mount an extension rod
carrying a motion-capture marker cluster beneath the camera
platform and another beneath the target vehicle.
Both marker clusters remain within the underwater
motion-capture system's observable volume throughout the
transitions.
Known fixed extrinsic transformations map the measured
marker-body poses to the poses of the camera optical frame
and target body frame, respectively.
These poses are synchronized with the recorded images to
support continuous ground-truth construction for the
22,346-frame cross-medium test set.

\begin{figure}[t]
  \centering
  \includegraphics[width=\columnwidth]{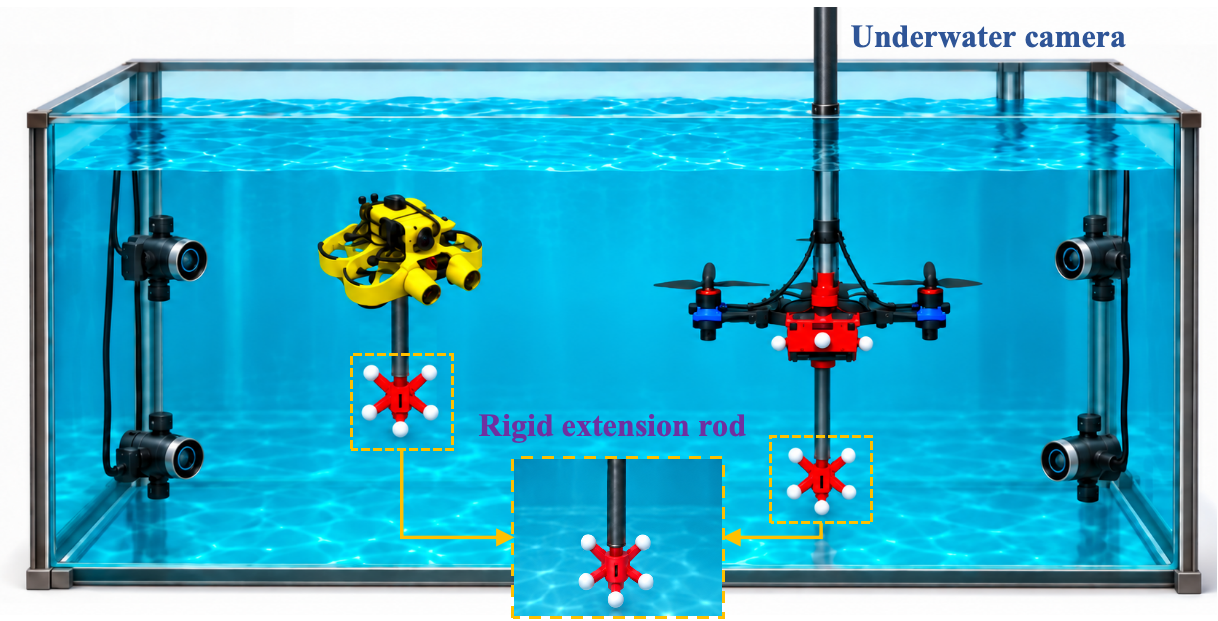}
  \caption{Underwater motion-capture setup for continuous
  camera and target pose measurement during air--water transitions.}
  \label{fig:mocap_setup}
\end{figure}

\subsection{Temporal Synchronization and Coordinate Projection}
Let $\mathcal F_w$, $\mathcal F_c$, $\mathcal F_o$, and $\mathcal F_g$ denote the motion-capture world frame, camera optical frame, target rigid-body frame, and target geometry frame, respectively. Camera images and motion-capture poses are recorded with the same monotonic clock and associated by nearest timestamp. A reference point $p_r^g$ and geometry vertices $q_j^g$ are transformed into the camera frame as
\begin{equation}
  p_{r,t}^{c} = T_{w,t}^{c}T_{o,t}^{w}T_g^o p_r^g,
  \quad
  q_{j,t}^{c} = T_{w,t}^{c}T_{o,t}^{w}T_g^o q_j^g.
  \label{eq:transform}
\end{equation}
The transforms act on homogeneous point coordinates~\cite{hartley2004multiple}. The calibrated camera model $\pi(\cdot)$ gives the initial reference-point projection $\tilde u_{r,t}=\pi(p_{r,t}^{c})$ and the initial image projections of the target geometry. This chain transfers the measured six-degree-of-freedom pose into the image without using target appearance.

\subsection{Medium-Gated Residual and Geometry-Based Box}
A standard pinhole projection does not fully describe the systematic shift introduced by the practical underwater imaging chain. We therefore use a lightweight model $f_\phi$ to predict a two-dimensional reference-point residual:
\begin{equation}
  u_{r,t} = \tilde u_{r,t} + g_t f_\phi(z_{r,t}),
  \label{eq:residual}
\end{equation}
where $g_t=1$ when the reference point is underwater and $g_t=0$ in air. The feature vector $z_{r,t}$ contains only calibrated geometric variables, including the three-dimensional coordinates and initial projection; it contains no target crop or tracking result. Consequently, the residual model corrects the imaging geometry rather than detecting the target.

Projecting the geometry vertices produces an initial box $\tilde B_t=(x_{\min},y_{\min},x_{\max},y_{\max})$. The gated residual $(\Delta u_t,\Delta v_t)=g_t f_\phi(z_{r,t})$ translates the entire box to form an annotation candidate:
\begin{equation}
  \hat B_t=\tilde B_t+[\Delta u_t,\Delta v_t,\Delta u_t,\Delta v_t].
  \label{eq:bbox}
\end{equation}
Thus, the learned model corrects candidate position only; width and height remain determined by known geometry and measured pose. This shared-translation approximation does not explicitly model differential refractive displacement across the target. The final box $B_t^*$ is obtained after manual review.

\subsection{Manual Review and Evaluation Set}
The projected boxes are treated as annotation candidates rather than accepted automatically. Reviewers inspect the synchronized image, projected box, target trajectory, and frame-to-frame consistency, then correct misalignment and flag invalid measurements. The final data retain the camera images, continuous two-dimensional boxes, synchronized pose provenance, and quality-control records.

The resulting evaluation-only cross-medium test set contains {\crosssequences} videos with {\crossframes} frames recorded at $60$~FPS, focusing on the stages surrounding air--water transitions. It is not used to train or fine-tune any tracker. All compared methods receive the same image sequences and first-frame target boxes; their predictions are evaluated against the same reviewed annotations. The set supports analysis of localization, drift during interface-induced degradation, and stable recovery following tracking loss.

%% file: sections/05_tracking_method.tex
\section{{\methodname}: Recovery-Oriented Tracking}

{\methodname} is an inference-time strategy for multi-template trackers, described here using MixFormerV2 without backbone retraining. Given image $I_t$ and the current appearance reference, the tracker produces a visual box $B_t^v$ and confidence $q_t^v$. Confidence-triggered template selection is the central design; motion feedback, search adaptation, and recovery verification provide supporting control.

\subsection{Reliability-Gated Motion Feedback}
We use an eight-dimensional constant-velocity Kalman filter~\cite{kalman1960} to represent target center, scale, and velocity:
\begin{equation}
  x_t=[c_x,c_y,w,h,\dot c_x,\dot c_y,\dot w,\dot h]^\top.
  \label{eq:state}
\end{equation}
The prediction step produces a motion box $B_t^m$ and covariance $P_t^-$. For the visual measurement $z_t^v$, we compute the innovation and its Mahalanobis distance,
\begin{equation}
  d_t^2=(z_t^v-Hx_t^-)^\top
  (HP_t^-H^\top+R_t)^{-1}(z_t^v-Hx_t^-).
  \label{eq:innovation}
\end{equation}
where $H$ maps the motion state to the box measurement and $R_t$ denotes visual measurement uncertainty.
A visual result is accepted only when its model confidence, innovation distance, normalized center displacement, and logarithmic scale change jointly pass their gates. Writing this decision as $a_t\in\{0,1\}$, the posterior is
\begin{equation}
  x_t=
  \begin{cases}
    \operatorname{KFUpdate}(x_t^-,z_t^v), & a_t=1,\\
    x_t^-, & a_t=0.
  \end{cases}
  \label{eq:gated_update}
\end{equation}
The confidence gate tests appearance evidence, while the remaining gates check temporal consistency.

\subsection{Uncertainty-Adaptive Search}
Search is centered on the motion prediction. Its factor increases with position uncertainty and the duration $\ell_t$ of unreliable tracking:
\begin{equation}
  s_t=\operatorname{clip}\!\left(
  s_0+\lambda_p\frac{\sqrt{P^-_{c_xc_x}+P^-_{c_yc_y}}}
  {\sqrt{w_t^-h_t^-}}+\lambda_\ell\ell_t,
  s_{\min},s_{\max}\right).
  \label{eq:adaptive_search}
\end{equation}
The bounds limit search expansion during prolonged unreliable tracking.

If local search remains unreliable, a wider redetection stage ranks candidates $b_i$ using visual confidence $q_i$, motion consistency $m_i$, and similarity $h_i$ to the retained appearance references:
\begin{equation}
  r_i=\alpha q_i+\beta m_i+\gamma h_i,
  \qquad \alpha+\beta+\gamma=1.
  \label{eq:redetection_score}
\end{equation}
The highest-ranked candidate must remain reliable for multiple consecutive frames before stable tracking is restored. This localization check does not promote the candidate into a new template.

\subsection{Confidence-Triggered Template Update}
\label{sec:template_policy}
At the first low-confidence frame $t_e$, where $q_{t_e}^v<\theta_{\mathrm{low}}$, we select a window-best historical template from the preceding $W$-frame window $\mathcal W_e$, excluding $t_e$:
\begin{equation}
  j_e=\operatorname*{arg\,max}_{j\in\mathcal W_e}q_j^v.
  \label{eq:window_selection}
\end{equation}
Let $\mathcal C(I,B)$ denote the template crop from image $I$ at box $B$, and let $B_{t_e}^m$ be the Kalman-predicted box for the trigger frame. The three template sources are
\begin{equation}
\begin{aligned}
  Z_0 &= \mathcal C(I_1,B_1),\\
  Z_{\mathrm{safe}}^{(e)} &= \mathcal C(I_{j_e},B_{j_e}^v),\\
  Z_{\mathrm{KF}}^{(e)} &= \mathcal C(I_{t_e},B_{t_e}^m).
\end{aligned}
  \label{eq:templates}
\end{equation}
The initial template anchors target identity, and the window-best historical template provides a pre-trigger appearance reference. The KF template provides an auxiliary observation of the predicted target region under the occlusion and degradation present at the trigger frame; it is not assumed to be a clean target view. The template policy sets the three template inputs and their associated weights for the underlying tracker. The templates selected at $t_e$ are used starting from frame $t_e+1$. The initial template remains fixed throughout the sequence. During the same low-confidence episode, the two selected templates are retained while Kalman prediction continues to provide position guidance.

\subsection{Scale Stability}

Medium transitions can also produce implausible box-scale jumps. We apply a weak soft scale constraint:
\begin{equation}
  (w_t,h_t)\leftarrow(1-\rho)(w_t^v,h_t^v)+\rho(\bar w_t,\bar h_t),
  \quad \rho=0.07,
  \label{eq:scale}
\end{equation}
where $(\bar w_t,\bar h_t)$ is estimated from recent reliable tracking history. The small weight suppresses abnormal feedback while retaining legitimate scale variation.The final output $B_t$ follows the visual estimate when reliable and falls back to the Kalman-predicted box $B_t^m$ when visual confidence remains insufficient.

%% file: sections/06_experiments.tex
\section{Experiments}

Our experiments examine projection accuracy for label construction, cross-medium tracking accuracy and recovery, and the effect of {\methodname} on conventional tracking benchmarks. No motion-capture measurement is available to any tracker during these experiments.

\subsection{Experimental Setup and Metrics}
The projection study uses {\projectionpoints} physically valid ChArUco control points from $722$ fixed validation images. MixFormerV2 comparisons use a fixed MixFormerV2-Base checkpoint and common first-frame initialization. Static performs no online replacement; Official selects the highest-confidence candidate within each update window; Fixed replaces a template periodically without gating; and Event commits the first qualified candidate after a cooldown interval. Event denotes this baseline, not the proposed low-confidence-triggered policy. MixFormerV2 is abbreviated as MFV2 in the tables. The same {\methodname} module and strategy are applied to MixFormerV2 and ODTrack. OSTrack is evaluated without {\methodname} because the implementation used here does not support multiple templates.The same core {\methodname} parameters are used for the accuracy and recovery evaluations.

Standard tracking accuracy is evaluated on all {\crosssequences} cross-medium videos ({\crossframes} frames) at the original image resolution. We report Macro Success AUC as the primary standard metric, together with precision, normalized precision, OP50, and OP75 when available. Macro Success AUC is the equal-weight mean of the sequence-level Success AUC values, preventing longer recordings from dominating the result. Tracking coverage is evaluated on the same 9,166 selected
frames for all methods, while recovery latency follows
the protocol in Section~\ref{sec:problem}. The LaSOT test set contains $280$ sequences and $685{,}360$ frames, and UAV123 contains $123$ sequences and $112{,}578$ frames. The cross-medium set is not used for training or fine-tuning.

\subsection{Projection Accuracy of Label Construction}
All projection variants use the same valid three-dimensional control points and independent two-dimensional image detections. Because the fixed validation set is also used to select the stored MLP checkpoint, this result verifies the construction mechanism rather than claiming generalization to an independent projection test set.

\begin{table}[t]
\centering
\caption{Motion-capture projection error (px).}
\label{tab:projection}
\resizebox{\columnwidth}{!}{%
\begin{tabular}{lrrrr}
\toprule
Method & Mean & Median & P95 & Max \\
\midrule
Geometric projection & 39.68 & 27.81 & 115.04 & 204.84 \\
Geometric + fixed offset & 39.68 & 30.96 & 111.64 & 210.47 \\
Geometric + linear residual & 20.20 & 17.03 & 46.83 & 96.29 \\
Planar-refraction diagnostic & 110.68 & 93.39 & 258.65 & 709.66 \\
\best{Geometric + MLP residual} & \best{3.00} & \best{2.33} & \best{7.67} & \best{47.55} \\
\bottomrule
\end{tabular}}
\end{table}

Relative to direct geometric projection, the medium-gated MLP residual reduces median error by $91.6\%$ and P95 error by $93.3\%$. A linear residual explains part of the systematic displacement but retains a pronounced long tail. The current planar-refraction model reaches its parameter boundary and is retained only as a diagnostic baseline. The corrected projection supplies annotation candidates for final manual review.

\subsection{Controllable Cross-Medium Image Acquisition Platform}
\label{sec:acquisition_platform}

\begin{figure}[t]
  \centering
  \includegraphics[width=\columnwidth]{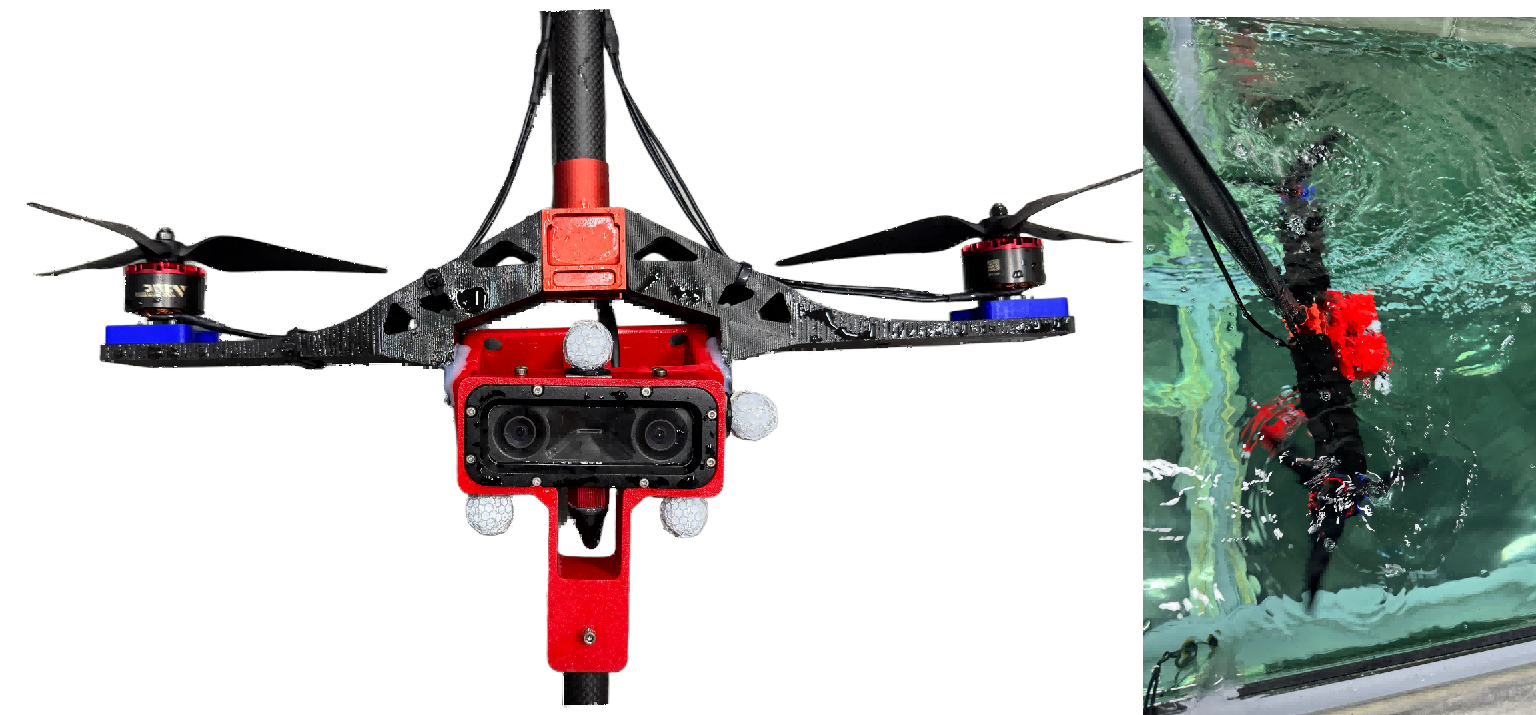}
  \caption{Controllable cross-medium image acquisition platform
  with an underwater camera and two independently driven
  propellers for generating bubbles, splashes, and
  water-surface disturbances.}
  \label{fig:acquisition_platform}
\end{figure}

To reproduce visual disturbances associated with water takeoff and entry under controlled conditions, we built a platform with an underwater camera and two independently driven propellers (Fig.~\ref{fig:acquisition_platform}). The camera is rigidly mounted at the lower end of a carbon-fiber support rod, with the propellers above it. Both the camera platform and the target move during recording, simulating tracking from a camera mounted on a cross-medium vehicle.
When operated near the surface, the propellers generate bubbles, splashes, and surface fluctuations that occlude the target and distort its appearance. Adjusting their operating conditions varies disturbance intensity, from mild interference to severe occlusion. This controllable setup supports efficient data collection and the study of tracking and recovery under interface-induced disturbances.

The motion-capture setup is described in
Section~\ref{sec:mocap_setup}.

\subsection{Template-Update Behavior Across Domains}
Before evaluating {\methodname}, we examine whether ordinary template-update rules transfer across data domains. All entries use the same MixFormerV2 backbone and a common inference protocol.

\begin{table}[t]
\centering
\caption{Success AUC of template-update strategies across data domains.}
\label{tab:template}
\small
\begin{tabular}{lrrr}
\toprule
Strategy & LaSOT & UAV123 & Cross-medium \\
\midrule
Static & 68.95 & \best{69.72} & \best{47.31} \\
Official & \best{70.26} & 69.60 & 46.95 \\
Fixed & 67.49 & 68.70 & 44.90 \\
Event & 68.36 & 69.25 & 46.04 \\
\bottomrule
\end{tabular}
\end{table}

Fixed updating yields the lowest AUC in all three domains. Official updating improves LaSOT AUC by $1.31$ points over Static, but decreases cross-medium AUC by $0.36$ points. These results motivate adapting template-update timing to degraded observations.

\begin{figure*}[!t]
  \centering
  \includegraphics[
    width=0.85\textwidth,
    height=\textheight,
    keepaspectratio
  ]{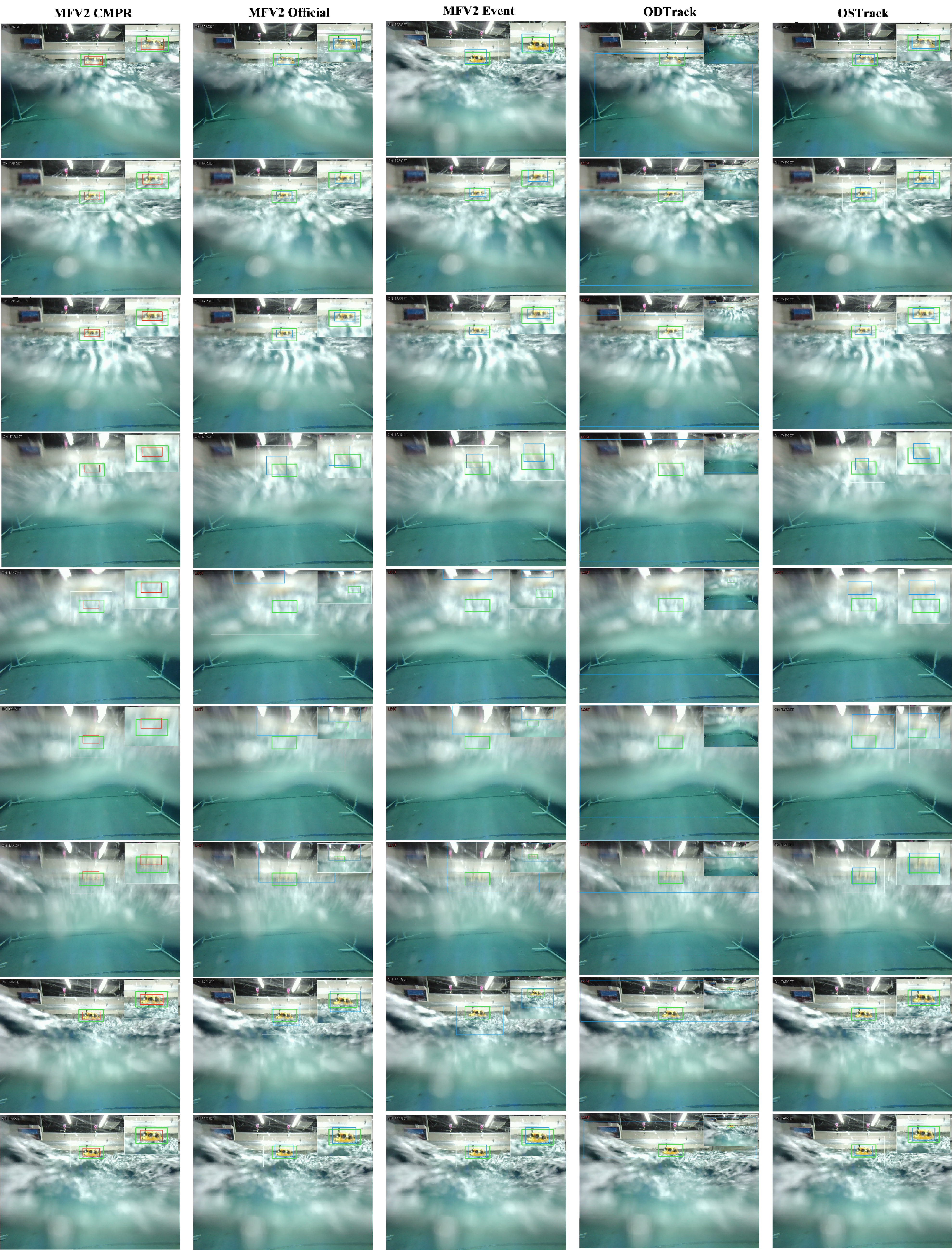}
  \caption{
    Qualitative comparison on representative frames from
    the cross-medium test set.
    From left to right: MixFormerV2 + {\methodname} (ours),
    MixFormerV2 with the Event template-update policy,
    MixFormerV2 with the Official template-update policy,
    ODTrack, and OSTrack.
    Green boxes indicate the ground truth (GT);
    the other colored boxes indicate tracker predictions.
  }
  \label{fig:qualitative_comparison}
\end{figure*}

\subsection{Standard Cross-Medium Tracking Accuracy}
Table~\ref{tab:overall} compares OSTrack~\cite{ye2022ostrack}, ODTrack~\cite{zheng2024odtrack}, MixFormerV2 Official, and the policy-augmented variants on the complete cross-medium test set.

\begin{table}[t]
\centering
\caption{Standard tracking metrics on the complete cross-medium test set.}
\label{tab:overall}
\small
\begin{tabular}{lrrrrr}
\toprule
Method & \shortstack{Macro\\AUC} & Prec. & \shortstack{Norm.\\Prec.} & OP50 & OP75 \\
\midrule
OSTrack & 39.03 & 41.18 & 54.28 & 48.20 & 14.10 \\
ODTrack & 29.77 & 27.18 & 36.36 & 32.51 & 11.64 \\
ODTrack + {\methodname} & 31.29 & 27.18 & 39.54 & 39.51 & \best{16.51} \\
MFV2 Official & 46.95 & 53.19 & \best{68.41} & 59.27 & 13.24 \\
MFV2 + {\methodname} & \best{49.90} & \best{53.30} & 68.40 & \best{67.40} & 15.10 \\
\bottomrule
\end{tabular}
\end{table}

MixFormerV2 with {\methodname} achieves $49.90$ Macro Success AUC, improving over Official updating by $2.95$ points and over Static in Table~\ref{tab:template} by $2.59$ points. OP50 increases from $59.27$ to $67.40$, while center-based precision remains nearly unchanged, indicating that the gains are more evident in overlap-based metrics. The ODTrack policy variant improves AUC from $29.77$ to $31.29$. Fig.~\ref{fig:qualitative_comparison} provides a qualitative comparison.

\subsection{Tracking Coverage and Recovery}

Table~\ref{tab:recovery} reports the number and percentage
of successfully tracked frames over the selected intervals,
together with mean video-level recovery latency following
Section~\ref{sec:problem}.
The last column reports the latency reduction of
MixFormerV2 + {\methodname} relative to each row method.

\begin{table}[t]
  \centering
  \setlength{\abovecaptionskip}{3pt}
  \caption{Tracking coverage on 9,166 selected frames and recovery latency.}
  \label{tab:recovery}
  \footnotesize
  \setlength{\tabcolsep}{2.5pt}
  \renewcommand{\arraystretch}{1.10}

  \begin{tabular*}{\columnwidth}{@{\extracolsep{\fill}}lrrr@{}}
    \toprule
    Method
    & \shortstack{Tracked\\frames $\uparrow$}
    & \shortstack{Mean\\frames $\downarrow$}
    & \shortstack{Latency\\reduction} \\
    \midrule
    OSTrack
      & 3725/9166 (40.64\%) & 63.0 & 21.7\% \\
    ODTrack
      & 1730/9166 (18.87\%) & 159.0 & 69.0\% \\
    ODTrack + {\methodname}
      & 1918/9166 (20.93\%) & 112.0 & 56.0\% \\
    MFV2 Static
      & 4478/9166 (48.85\%) & 54.8 & 10.0\% \\
    MFV2 Official
      & 4391/9166 (47.91\%) & 55.3 & 10.8\% \\
    MFV2 Event
      & 4294/9166 (46.85\%) & 59.6 & 17.3\% \\
    MFV2 Fixed
      & 3798/9166 (41.44\%) & 65.0 & 24.2\% \\
    MFV2 + {\methodname}
      & \textbf{4622/9166 (50.43\%)}
      & \textbf{49.3} & Ref. \\
    \bottomrule
  \end{tabular*}
\end{table}

With {\methodname}, MixFormerV2 increases the number of
successfully tracked frames from $4{,}391$ to $4{,}622$
($47.91\%$ to $50.43\%$ coverage), compared with Official updating.
Mean loss-to-recovery latency over each method's successfully
recovered videos decreases from $55.3$ to $49.3$ frames
($10.8\%$).
For ODTrack, {\methodname} increases the tracked-frame count
from $1{,}730$ to $1{,}918$ ($18.87\%$ to $20.93\%$ coverage)
and reduces mean recovery latency from $159$ to $112$ frames
($29.6\%$).

\subsection{Stability on Conventional Tracking Data}
Finally, Table~\ref{tab:general} evaluates the policy on LaSOT and UAV123 to characterize its effect on conventional tracking.

\begin{table}[t]
\centering
\caption{Tracking performance on conventional benchmarks.}
\label{tab:general}
\footnotesize
\begin{tabular}{llrrrrr}
\toprule
Dataset & Method & AUC & Prec. & \shortstack{Norm.\\Prec.} & OP50 & OP75 \\
\midrule
LaSOT & MFV2 Official & 70.26 & 75.83 & 80.09 & 82.05 & 69.13 \\
LaSOT & MFV2 + {\methodname} & 69.91 & 75.41 & 79.63 & 81.58 & 68.72 \\
UAV123 & MFV2 Official & 69.60 & 90.71 & 85.23 & 85.28 & 63.09 \\
UAV123 & MFV2 + {\methodname} & 69.18 & 90.16 & 84.81 & 84.74 & 62.61 \\
\bottomrule
\end{tabular}
\end{table}

On LaSOT and UAV123, the policy decreases AUC by $0.35$ and $0.42$ points, respectively, indicating a small conventional-tracking trade-off.

\subsection{Towards Robotic Deployment}

\begin{figure}[t]
  \centering
  \includegraphics[width=\columnwidth]{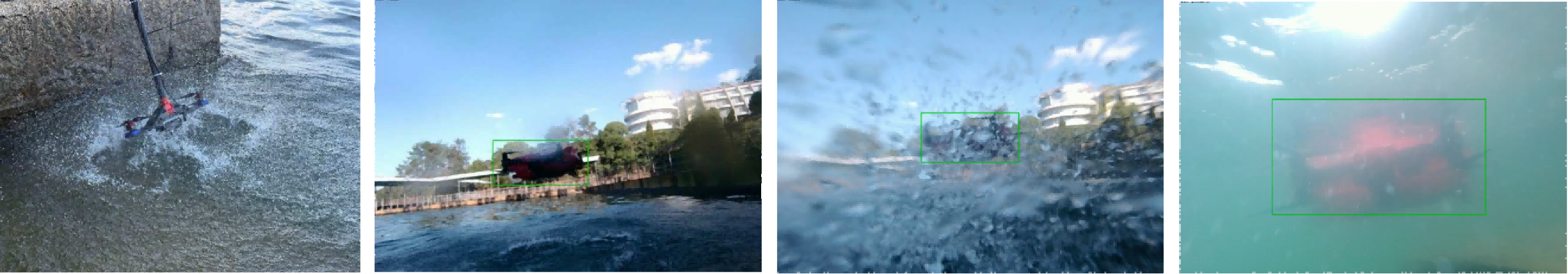}
  \caption{Outdoor acquisition setup and qualitative tracking results.}
  \label{fig:outdoor_results}
\end{figure}

To support future vision-based formation control of two
air--water vehicles, we conducted preliminary outdoor
experiments using the controllable image acquisition
platform described in Section~\ref{sec:acquisition_platform}.
The platform reproduced the visual conditions encountered
by an observing vehicle during air--water transitions while
recording a target vehicle entering and leaving the water.
Fig.~\ref{fig:outdoor_results} shows the outdoor acquisition
setup and representative qualitative tracking results
obtained by offline evaluation of the unannotated recordings.
Detailed qualitative tracking results are provided in the
accompanying supplementary video.

The complete MixFormerV2-based tracker with {\methodname}
was deployed on an NVIDIA Jetson Orin NX, processing
recorded video at over $27$~FPS. This result supports the
feasibility of embedded visual tracking for future
air--water robotic applications.

%% file: sections/07_conclusion.tex
\section{Conclusion}

We presented a framework for continuous evaluation and
recovery-oriented air--water tracking. Motion-capture-based
ground-truth construction provides an evaluation-only test
set of {\crosssequences} videos and {\crossframes} frames
with continuous physical-position references.
Without backbone retraining, {\methodname} combines
confidence-triggered template selection with motion guidance,
improving MixFormerV2 Macro Success AUC by $2.95$ points
over Official updating. Mean loss-to-recovery latency
decreases from $55.3$ to $49.3$ frames, computed over
each method's successfully recovered videos.
These gains involve small AUC decreases on LaSOT and UAV123.

Offline outdoor tests further illustrate tracking capability,
while recorded-video processing at over $27$~FPS on a
Jetson Orin NX supports embedded implementation.
Future work will broaden evaluation across targets and
environments and validate closed-loop visual formation
control of two air--water vehicles.

%% file: references.bib
@inproceedings{wu2013otb,
  author = {Wu, Yi and Lim, Jongwoo and Yang, Ming-Hsuan},
  title = {Online object tracking: A benchmark},
  booktitle = {Proc. IEEE CVPR},
  pages = {2411--2418}, year = {2013}
}

@inproceedings{ye2022ostrack,
  author = {Ye, Botao and Chang, Hong and Ma, Bingpeng and Shan, Shiguang and Chen, Xilin},
  title = {Joint feature learning and relation modeling for tracking: A one-stream framework},
  booktitle = {Proc. ECCV},
  pages = {341--357}, year = {2022}
}

@inproceedings{cui2022mixformer,
  author = {Cui, Yutao and Jiang, Cheng and Wang, Limin and Wu, Gangshan},
  title = {{MixFormer}: End-to-end tracking with iterative mixed attention},
  booktitle = {Proc. IEEE/CVF CVPR},
  pages = {13608--13618}, year = {2022}
}

@inproceedings{cui2023mixformerv2,
  author = {Cui, Yutao and Song, Tianhui and Wu, Gangshan and Wang, Limin},
  title = {{MixFormerV2}: Efficient fully transformer tracking},
  booktitle = {Proc. NeurIPS},
  pages = {58736--58751}, year = {2023}
}

@inproceedings{zheng2024odtrack,
  author = {Zheng, Yaozong and Zhong, Bineng and Liang, Qihua and Mo, Zhiyi and Zhang, Shengping and Li, Xianxian},
  title = {{ODTrack}: Online dense temporal token learning for visual tracking},
  booktitle = {Proc. AAAI},
  pages = {7588--7596}, year = {2024}
}

@inproceedings{akkaynak2019seathru,
  author = {Akkaynak, Derya and Treibitz, Tali},
  title = {{Sea-thru}: A method for removing water from underwater images},
  booktitle = {Proc. IEEE/CVF CVPR},
  pages = {1682--1691}, year = {2019}
}

@inproceedings{fan2019lasot,
  author = {Fan, Heng and Lin, Liting and Yang, Fan and Chu, Peng and Deng, Ge and Yu, Sijia and others},
  title = {{LaSOT}: A high-quality benchmark for large-scale single object tracking},
  booktitle = {Proc. IEEE/CVF CVPR},
  pages = {5374--5383}, year = {2019}
}

@inproceedings{mueller2016uav123,
  author = {Mueller, Matthias and Smith, Neil and Ghanem, Bernard},
  title = {A benchmark and simulator for {UAV} tracking},
  booktitle = {Proc. ECCV},
  pages = {445--461}, year = {2016}
}

@article{kalman1960,
  author = {Kalman, Rudolf E.},
  title = {A new approach to linear filtering and prediction problems},
  journal = {J. Basic Eng.},
  volume = {82}, number = {1}, pages = {35--45}, year = {1960}
}

@article{siddall2017aquamav,
  author = {Siddall, Robert and Ortega Ancel, Alejandro and Kova{\v{c}}, Mirko},
  title = {Wind and water tunnel testing of a morphing aquatic micro air vehicle},
  journal = {Interface Focus},
  volume = {7}, number = {1}, pages = {20160085}, year = {2017}
}

@article{chen2017aerialaquatic,
  author = {Chen, Yufeng and Wang, Hongqiang and Helbling, E. Farrell and Jafferis, Noah T. and Zufferey, Raphael and Ong, Aaron and others},
  title = {A biologically inspired, flapping-wing, hybrid aerial-aquatic microrobot},
  journal = {Sci. Robot.},
  volume = {2}, number = {11}, pages = {eaao5619}, year = {2017}
}

@article{li2022aerialaquatic,
  author = {Li, Lei and Wang, Siqi and Zhang, Yiyuan and Song, Shanyuan and Wang, Chuqian and Tan, Shaochang and others},
  title = {Aerial-aquatic robots capable of crossing the air-water boundary and hitchhiking on surfaces},
  journal = {Sci. Robot.},
  volume = {7}, number = {66}, pages = {eabm6695}, year = {2022}
}

@inproceedings{zhang2019updatenet,
  author = {Zhang, Lichao and Gonzalez-Garcia, Abel and van de Weijer, Joost and Danelljan, Martin and Khan, Fahad Shahbaz},
  title = {Learning the model update for {Siamese} trackers},
  booktitle = {Proc. IEEE/CVF ICCV},
  pages = {4010--4019}, year = {2019}
}

@article{an2023trusted,
  author = {An, Zhiyong and Chen, Geng and Zhang, Ximin and Wang, Zhuhai and Xie, Qingsong and Zhang, Bin},
  title = {Learning the model update with local trusted templates for visual tracking},
  journal = {IET Image Process.},
  volume = {17}, number = {2}, pages = {544--557}, year = {2023}
}

@article{yang2026samurai,
  author = {Yang, Cheng-Yen and Huang, Hsiang-Wei and Chai, Wenhao and Jiang, Zhongyu and Hwang, Jenq-Neng},
  title = {{SAMURAI}: Motion-aware memory for training-free visual object tracking with {SAM 2}},
  journal = {IEEE Trans. Image Process.},
  volume = {35}, pages = {970--982}, year = {2026}
}

@article{alzu2018looncopter,
  author = {Alzu'bi, Hamzeh and Mansour, Iyad and Rawashdeh, Osamah},
  title = {{Loon Copter}: Implementation of a hybrid unmanned aquatic--aerial quadcopter with active buoyancy control},
  journal = {J. Field Robot.},
  volume = {35}, number = {5}, pages = {764--778}, year = {2018}
}

@article{mercado2019aerialunderwater,
  author = {Mercado, D. and Maia, M. and Diez, F. J.},
  title = {Aerial-underwater systems, a new paradigm in unmanned vehicles},
  journal = {J. Intell. Robot. Syst.},
  volume = {95}, number = {1}, pages = {229--238}, year = {2019}
}

@article{zufferey2019jumpgliding,
  author = {Zufferey, R. and Ortega Ancel, A. and Farinha, A. and Siddall, R. and Armanini, S. F. and Nasr, M. and others},
  title = {Consecutive aquatic jump-gliding with water-reactive fuel},
  journal = {Sci. Robot.},
  volume = {4}, number = {34}, pages = {eaax7330}, year = {2019}
}

@inproceedings{rockenbauer2021dipper,
  author = {Rockenbauer, Friedrich M. and Jeger, Simon L. and Beltran, Liberto and Berger, Maximilian A. and Harms, Marvin and Kaufmann, Noah and others},
  title = {{Dipper}: A dynamically transitioning aerial-aquatic unmanned vehicle},
  booktitle = {Proc. Robot.: Sci. Syst. (RSS)},
  year = {2021}
}

@article{zufferey2026leaping,
  author = {Zufferey, Raphael and Jeger, Simon L. and H{\"u}sser, Moritz and Ruiz, Fernando and Lapsansky, Anthony and Ijspeert, Auke and others},
  title = {Leaping out of the water: Aerial-aquatic locomotion with flapping wings},
  journal = {Science},
  volume = {393}, number = {6807}, pages = {207--211}, year = {2026}
}

@inproceedings{bi2026useroriented,
  author = {Bi, Yuanbo and Wang, Youdong and Chen, Yikai and Yao, Kanzhong and Peng, Zimeng and Wang, Jiaguo and others},
  title = {A user-oriented multimodal {FPV} drone system: Design and implementation for aquatic-aerial operation},
  booktitle = {Proc. OCEANS},
  pages = {1--9}, year = {2026}
}

@inproceedings{alawode2023utb180,
  author = {Alawode, Basit and Guo, Yuhang and Ummar, Mehnaz and Werghi, Naoufel and Dias, Jorge and Mian, Ajmal and others},
  title = {{UTB180}: A high-quality benchmark for underwater tracking},
  booktitle = {Proc. ACCV},
  pages = {442--458}, year = {2023}
}

@inproceedings{zhang2024webuot1m,
  author = {Zhang, Chunhui and Liu, Li and Huang, Guanjie and Wen, Hao and Zhou, Xi and Wang, Yanfeng},
  title = {{WebUOT-1M}: Advancing deep underwater object tracking with a million-scale benchmark},
  booktitle = {Proc. NeurIPS},
  pages = {50152--50167}, year = {2024}
}

@inproceedings{dai2020ltmu,
  author = {Dai, Kenan and Zhang, Yunhua and Wang, Dong and Li, Jianhua and Lu, Huchuan and Yang, Xiaoyun},
  title = {High-performance long-term tracking with meta-updater},
  booktitle = {Proc. IEEE/CVF CVPR},
  pages = {6298--6307}, year = {2020}
}

@inproceedings{valmadre2018oxuva,
  author = {Valmadre, Jack and Bertinetto, Luca and Henriques, Jo{\~a}o F. and Tao, Ran and Vedaldi, Andrea and Smeulders, Arnold W. M. and others},
  title = {Long-term tracking in the wild: A benchmark},
  booktitle = {Proc. ECCV},
  pages = {692--707}, year = {2018}
}

@article{lukezic2021longterm,
  author = {Luke\v{z}i\v{c}, Alan and \v{C}ehovin Zajc, Luka and Voj{\'i}\v{r}, Tom{\'a}\v{s} and Matas, Ji\v{r}{\'i} and Kristan, Matej},
  title = {Performance evaluation methodology for long-term single-object tracking},
  journal = {IEEE Trans. Cybern.},
  volume = {51}, number = {12}, pages = {6305--6318}, year = {2021}
}

@article{burri2016euroc,
  author = {Burri, Michael and Nikolic, Janosch and Gohl, Pascal and Schneider, Thomas and Rehder, Joern and Omari, Sammy and others},
  title = {The {EuRoC} micro aerial vehicle datasets},
  journal = {Int. J. Robot. Res.},
  volume = {35}, number = {10}, pages = {1157--1163}, year = {2016}
}

@book{hartley2004multiple,
  author = {Hartley, Richard and Zisserman, Andrew},
  title = {Multiple View Geometry in Computer Vision},
  edition = {2}, publisher = {Cambridge Univ. Press}, year = {2004}
}

@inproceedings{mitrokhin2019evimo,
  author = {Mitrokhin, Anton and Ye, Chengxi and Ferm{\"u}ller, Cornelia and Aloimonos, Yiannis and Delbruck, Tobi},
  title = {{EV-IMO}: Motion segmentation dataset and learning pipeline for event cameras},
  booktitle = {Proc. IEEE/RSJ IROS},
  pages = {6105--6112}, year = {2019}
}

@article{mueggler2017eventdataset,
  author = {Mueggler, Elias and Rebecq, Henri and Gallego, Guillermo and Delbruck, Tobi and Scaramuzza, Davide},
  title = {The event-camera dataset and simulator: Event-based data for pose estimation, visual odometry, and {SLAM}},
  journal = {Int. J. Robot. Res.},
  volume = {36}, number = {2}, pages = {142--149}, year = {2017}
}

@inproceedings{schubert2018tumvi,
  author = {Schubert, David and Goll, Thore and Demmel, Nikolaus and Usenko, Vladyslav and St{\"u}ckler, J{\"o}rg and Cremers, Daniel},
  title = {The {TUM VI} benchmark for evaluating visual-inertial odometry},
  booktitle = {Proc. IEEE/RSJ IROS},
  pages = {1680--1687}, year = {2018}
}
